\documentclass[10pt, a4paper]{article}

\usepackage{microtype}
\usepackage{graphicx}
\usepackage{subfigure}
\usepackage{dirtytalk}
\usepackage{booktabs} 
\usepackage[nolist]{acronym}
\usepackage[table]{xcolor}
\usepackage[font=small,skip=25pt]{caption}

\usepackage{hyperref}

\usepackage[review]{lrec-coling2024} 

\usepackage{amsmath}
\usepackage{amssymb}
\usepackage{mathtools}
\usepackage{amsthm}

\usepackage[capitalize,noabbrev]{cleveref}

\theoremstyle{plain}

\theoremstyle{definition}

\theoremstyle{remark}

\usepackage{enumitem}                       

\title{MultiLegalPile: A 689GB Multilingual Legal Corpus}

\name{Joel Niklaus, Veton Matoshi, Matthias Stürmer, Ilias Chalkidis, Daniel E. Ho} 

\address{University of Bern, Bern University of Applied Sciences, Bern University of Applied Sciences, University of Copenhagen, Stanford University\\
         author1@xxx.yy, author2@zzz.edu, author3@hhh.com, author4@hhh.com, author5@hhh.com}

\abstract{
Large, high-quality datasets are crucial for training \acp{LLM}. However, so far, there are few datasets available for specialized critical domains such as law and the available ones are often only for the English language. We curate and release \textsc{MultiLegalPile}, a 689GB corpus in 24 languages from 17 jurisdictions. The \textsc{MultiLegalPile} corpus, which includes diverse legal data sources with varying licenses, allows for pretraining NLP models under fair use, with more permissive licenses for the Eurlex Resources and Legal mC4 subsets. We pretrain two RoBERTa models and one Longformer multilingually, and 24 monolingual models on each of the language-specific subsets and evaluate them on LEXTREME. Additionally, we evaluate the English and multilingual models on LexGLUE. Our multilingual models set a new SotA on LEXTREME and our English models on LexGLUE. We release the dataset, the trained models, and all of the code under the most open possible licenses.
 \\ \newline  }
 
\begin{acronym}[UMLX]
    \acro{FSCS}{Federal Supreme Court of Switzerland}
    \acro{SCI}{Supreme Court of India}
    \acro{ECHR}{European Convention of Human Rights}
    \acro{ECtHR}{European Court of Human Rights}
    \acro{SCOTUS}{Supreme Court of the United States}
    \acro{SPC}{Supreme People's Court of China}
    \acro{SJP}{Swiss-Judgment-Prediction}
    \acro{ASO}{Almost Stochastic Order}
    \acro{ILDC}{Indian Legal Documents Corpus}
    
    \acro{US}{United States}
    \acro{EU}{European Union}

    \acro{NLP}{Natural Language Processing}
    \acro{ML}{Machine Learning}
    \acro{LJP}{Legal Judgment Prediction}
    \acro{SJP}{Swiss-Judgment-Prediction}
    \acro{PJP}{Plea Judgment Prediction}
    
    \acro{BERT}{Bidirectional Encoder Representations from Transformers}
    \acro{LSTM}{Long Short-Term Memory }
    \acro{GRU}{Gated Recurrent Unit}
    \acro{BiLSTM}{Bidirectional Long Short-Term Memory}
    \acro{CNN}{Convolutional Neural Networks}

    \acro{PLM}{pretrained Language Model}
    \acro{LLM}{Large Language Model}

    \acro{CLT}{Cross-Lingual Transfer}
    \acro{HRL}{high resource language}
    \acro{LRL}{low resource language}

    \acro{POS}{Part-of-Speech}
    
    \acro{SLTC}{Single Label Text Classification}
    \acro{MLTC}{Multi Label Text Classification}
    \acro{NER}{Named Entity Recognition}
    \acro{NLU}{Natural Language Understanding}

    \acro{GNB}{Gaussian Naive Bayes}
    \acro{DT}{Decision Tree}
    \acro{SVM}{Support-vector Machine}
    \acro{RF}{ Random Forest}
    \acro{XGBoost}{eXtreme Gradient Boosting}
    \acro{MLIR}{Multilingual Information Retrieval}
    
    \acro{BGE}{Leading Decision}
    \acro{FSCD}{Federal Supreme Court Decisons}
    \acro{FSCS}{Swiss Federal Supreme Court}
\end{acronym}

\begin{document}

\maketitleabstract





\section{Introduction}





Recent years have seen \acp{LLM} achieving remarkable progress, as demonstrated by their performance on various benchmarks such as SuperGLUE \cite{wang_superglue_2019}, MMLU \cite{hendrycks_measuring_2021}, and several human Exams~\cite{openai_gpt-4_2023}, including U.S. bar exams for admission to practice law~\cite{katz_gpt-4_2023}. These models are typically trained on increasingly large corpora, such as the Pile \cite{gao_pile_2020}, C4 \cite{raffel_exploring_2020}, and mC4 \cite{xue_mt5_2021}. However, public corpora available for training these models are predominantly in English, and often constitute web text with unclear licensing. This even led to lawsuits against \ac{LLM} producers\footnote{\tiny\url{https://www.theverge.com/2022/11/8/23446821/microsoft-openai-github-copilot-class-action-lawsuit-ai-copyright-violation-training-data}}, highlighting this critical issue.
\begin{figure}[ht]
    \centering
    \includegraphics[width=\columnwidth]{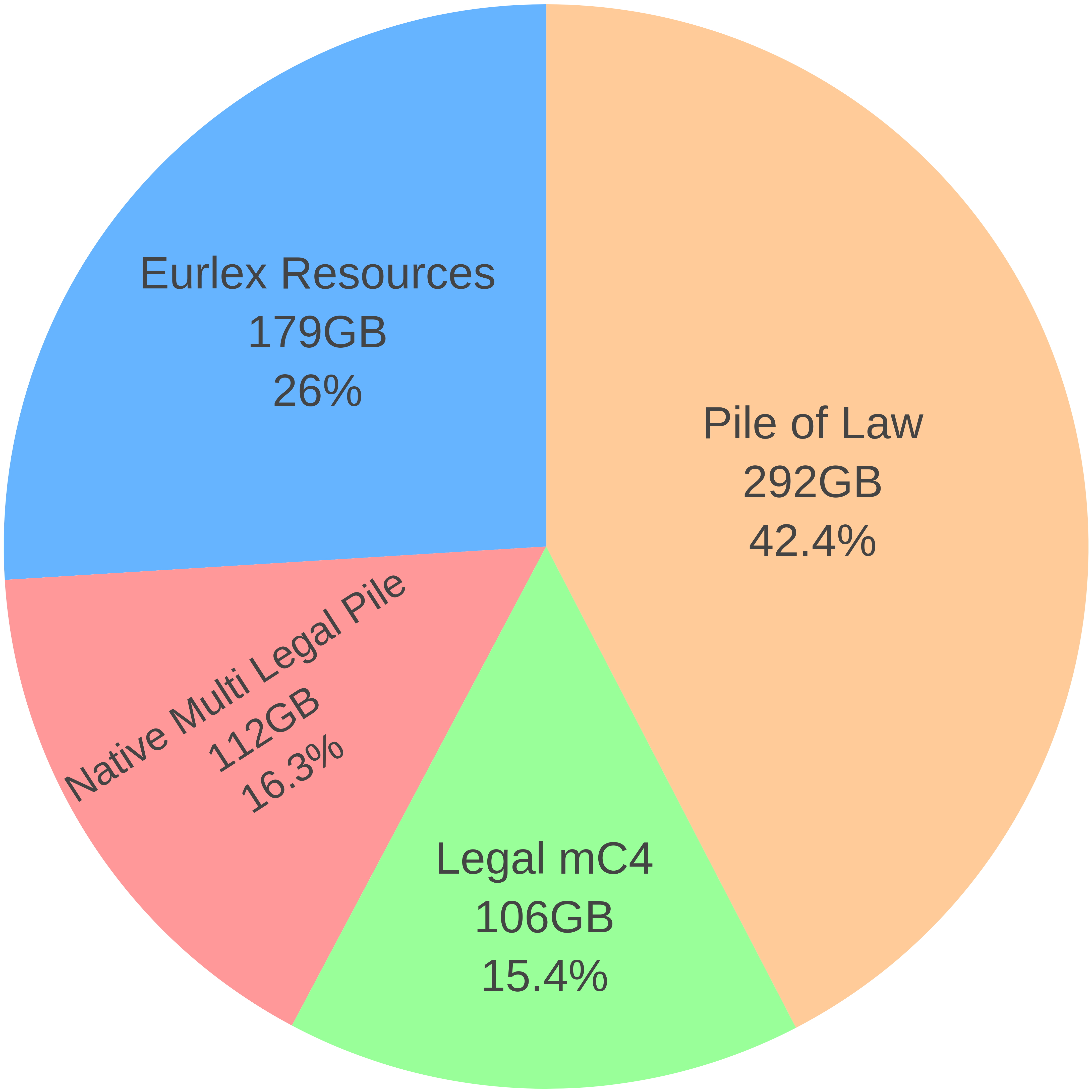}
    \caption{\textsc{MultiLegalPile} Source Distribution}
    \label{fig:source_distribution}
    \vspace{-5mm}
\end{figure}
Furthermore, there is a scarcity of large-scale, domain-specific pretraining corpora, which constitutes a significant gap in the current body of resources available for the training of \acp{LLM}. We find that only one in every thousand document in mC4 contains legal text. Similarly, \ac{LLM}s are predominantly English, especially considering domain-specific models, e.g., ones specialized in biomedical, legal, or financial texts.






Legal texts, often produced by public instruments (e.g., state governments, international organizations), are typically available under public licenses, offering a rich resource for domain-specific pretraining. Given this context, we curate a massive, openly available, corpus of multilingual law text spanning across numerous jurisdictions (legal systems), predominantly under permissive licenses.

Further on, we continue pretraining XLM-R models \cite{conneau_cross-lingual_2019} on our corpus and evaluated these models on the recently introduced LEXTREME \cite{niklaus_lextreme_2023} and LexGLUE \cite{chalkidis_lexglue_2022} benchmarks. 
Given the often extensive nature of legal text, we also pretrained a Longformer model \cite{beltagy_longformer_2020} for comparison with hierarchical models \cite{chalkidis-etal-2019-neural,niklaus_swiss-judgment-prediction_2021, niklaus_empirical_2022}.

Our multilingual models set a new state-of-the-art (SotA) on LEXTREME overall. Our legal Longformer outperforms all other models in four LEXTREME datasets and reaches the highest dataset aggregate score. Our monolingual models outperform their base model XLM-R in 21 out of 24 languages, even reaching language specific SotA in five. 
On LexGLUE our English models reach SotA in five out of seven tasks with the large model achieving the highest aggregate score. 

In the spirit of open science, we provide the dataset under a CC BY-NC-SA 4.0 license, with some subsets licensed more permissively. Dataset creation scripts, models, and pretraining code are public under Apache 2.0 licenses. This open-source approach encourages further research and advancements in the field of legal text analysis and understanding using \acp{LLM}.

\subsection*{Contributions}
The contributions of this paper are three-fold:
First, we curate and release a large multilingual legal text corpus, dubbed \textsc{MultiLegalPile},\footnote{
Link will be released upon acceptance.
} 
covering 24 languages and 17 legal systems (jurisdictions).
Second, we release two multilingual and 24 monolingual legal PLMs, termed \textsc{LegalXLMs}, initiated from XLM-R~\cite{conneau_cross-lingual_2019} and further pretrained on the \textsc{MultiLegalPile}. We also pretrain a Longformer \cite{beltagy_longformer_2020} based on our multilingual base-size model on context lengths of up to 4096 tokens.
Third, we benchmark the newly released models on LEXTREME and LexGLUE, reaching SotA for base- and large-size models and increasing performance drastically in Greek legal code. Our Longformer model achieves SotA in four tasks and the highest dataset aggregate score. Our monolingual models set language-specific SotA in five languages.

\section{Related Work}

In this section, we briefly discuss prior general and domain-specific pretraining corpora. See Appendix \ref{sec:additional_related_work} for a more detailed discussion of related works.

\subsection{General Pretraining Corpora}

The One Billion Word Language Model Benchmark (LM1B) \cite{chelba2014billion}, Wikipedia, and derived datasets like WikiText \cite{merity2016pointer} and BookCorpus \cite{BookCorpus2015} have been crucial in developing language models such as GPT-2 \cite{radford_language_2019}, BERT \cite{devlin-etal-2019-bert}, and RoBERTa \cite{liu_roberta_2019}. Large-scale datasets like the Colossal Clean Crawled Corpus (C4) \cite{raffel_exploring_2020}, OpenWebText \cite{Gokaslan2019OpenWeb}, The Pile \cite{ThePile2020}, and Glot500 \cite{imanigooghari2023glot500} have further advanced the field, contributing to the training of models like T5, MegatronBERT \cite{shoeybi2020megatronlm}, GPT-3 \cite{brown-etal-gpt3}, and Glot500-m. 
Although general pretraining datasets are large and widely available, we find that mC4 only contains around 0.1\% legal text (see Section \ref{sec:filtering_mc4}), exemplifying the need for datasets specifically tailored to the legal domain. 


\subsection{Domain Specific Corpora}

\begin{table}[ht]
    \centering
    \footnotesize
    \resizebox{\columnwidth}{!}{
    \begin{tabular}{llll}
         \toprule
         \textbf{Model} & \textbf{Domain} & \textbf{Languages} & \textbf{Size in \# Words} \\
         \midrule  
        SciBERT \cite{Beltagy2019SciBERTAP}         & scientific    & English       & 2.38B (3.17B tokens)\\
        Galactica \cite{taylor_galactica_2022}      & scientific    & English       & 79.5B (106B tokens)\\
        BioBERT \cite{Lee2020}             & biomedical    & English       & 18B\\
        LegalBERT \cite{chalkidis-etal-2020-legal}  & legal         & English       & 1.44B (11.5GB)\\ 
        CaselawBERT \cite{10.1145/3462757.3466088}  & legal         & English       & 4.63B (37GB)\\
        LexFiles \cite{chalkidis-etal-2020-legal}   & legal         & English       & 18.8B \\
        LegalXLMs (ours)                            & legal         & 24 EU langs   & 87B (689GB)\\
        \bottomrule
    \end{tabular}
    }
    \caption{Previous domain-specific pretraining corpora. For some, only GB or tokens were available. We converted 8 GB into 1B words and 1 token to 0.75 words.}
    \label{tab:domain_specific_corpora}
    \vspace{-5mm}
\end{table}

Pretraining on domain-specific text like medicine, law, or science can boost \ac{LM} performance on related tasks \cite{Beltagy2019SciBERTAP, Gu2021, chalkidis-etal-2020-legal, niklaus_budgetlongformer_2022}.
In the scientific field, SciBERT was pretrained on a mix of computer science and biomedical papers \cite{Beltagy2019SciBERTAP}. Similarly, models like PubMedBERT \cite{Gu2021} and BioBERT \cite{Lee2020} were pretrained on biomedical datasets. ClinicalBERT utilized the Medical Information Mart for Intensive Care III (MIMIC-III) dataset, encompassing 2 million clinical notes, demonstrating superior performance on medical NLP tasks \cite{Huang2019}.
In the legal realm, LegalBERT was pretrained on 12 GB of English legal texts, achieving high performance on domain-specific tasks \cite{chalkidis-etal-2020-legal}. CaseLaw-BERT utilized the English Harvard Law case corpus from 1965 to 2021 \cite{10.1145/3462757.3466088}. Recently, LexFiles was released, with 11 sub-corpora covering six English-speaking legal systems and 19B tokens \cite{chalkidis2023lexfiles}. It was used to train new legal English PLMs, showing enhanced results in legal tasks.
Though efforts to pretrain legal \acp{LM} exist in languages like Italian, Romanian, and Spanish \cite{licari_italian-legal-bert_2022, masala-etal-2021-jurbert, gutierrez-fandino_spanish_2021}, English remains predominant, emphasizing the need for multilingual legal corpora.
Table \ref{tab:domain_specific_corpora} compares previous domain-specific corpora, all in English and all legal corpora less than 1/4 of \textsc{MultiLegalPile}'s size.

\section{\textsc{MultiLegalPile}}
\label{sec:multilegalpile}
\subsection{Construction}


We transformed all datasets into xz compressed JSON Lines (JSONL) format. The combination of XZ compression and JSONL is ideal for streaming large datasets due to reduced file size and efficient decompression and reading.

\paragraph{Filtering mC4}
\label{sec:filtering_mc4}
We used the vast multilingual web crawl corpus, mC4 \cite{xue_mt5_2021}, as our base dataset. To effectively isolate legal content, we used regular expressions to identify documents with legal references, such as ``Art. 5'' or ``§ 8'' . We found that detecting legal citations, such as references to laws and rulings, served as a reliable indicator of legal-specific documents in the corpus.

\begin{table}[ht]
    \centering
    \footnotesize
    \resizebox{\columnwidth}{!}{
    \begin{tabular}{lrrrrr}
         \toprule
         \textbf{Iteration} & \textbf{German} & \textbf{English} & \textbf{Spanish} & \textbf{French} & \textbf{Italian} \\
         \midrule  
            1st & 100\% & 20\% & 100\% & 65\% & 80\%\\
            2nd & 100\% & 85\% & 100\% & 100\% & 95\%\\
        \bottomrule
    \end{tabular}
    }
    \caption{Per language precision in legal mC4 (n=20)}
    \label{tab:legal_mc4_precision}
\end{table}

To ensure the accuracy of our filtering, we engaged legal experts to aid in identifying citations to laws and rulings across different jurisdictions and languages. We manually reviewed the precision of the retrieved documents for five languages, namely German, English, Spanish, French, and Italian, as shown in Table \ref{tab:legal_mc4_precision}. The proficiency levels of the evaluators included native German, fluent English and Spanish, intermediate French, and basic Italian.

Subsequent to the initial review, we performed a second round of precision evaluation, during which we refined our regex expressions based on our findings from the first iteration. This iterative process not only enhanced the precision of the legal content detection, but also resulted in a reduction of the corpus size from 133GB to 106GB. Although the overall volume of data was reduced, this process significantly improved the quality and specificity of the corpus by focusing on legal content with a higher degree of precision.
A major reason for using regexes instead of an machine learning based classifier was speed. Already when utilizing regexes, filtering through such a huge corpus like mC4 (27TB in total, of which 10.4TB are in English) took several days on our hardware. An ML model based on Bag-of-Words, Word vectors or even contextualized embeddings would a) need an annotated dataset and b) likely be much slower.

\begin{figure*}[ht!]
    \centering
    \includegraphics[width=\textwidth]{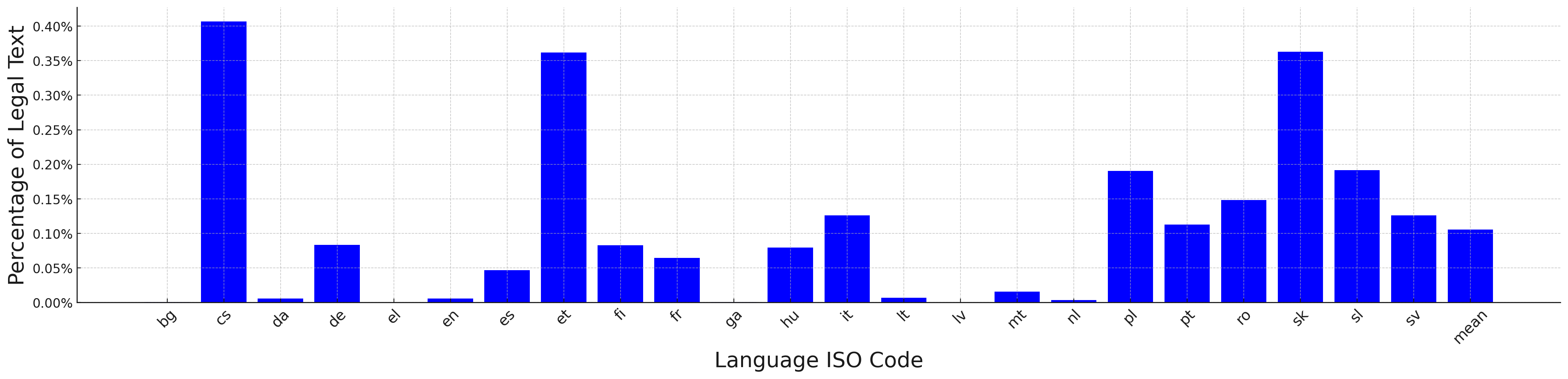}
    \caption{Percentage of Legal Text in mC4 per Language}
    \label{fig:percentage_legal}
\end{figure*}

We find that on average, only one in every thousand pages in mC4 contains legal content. We show a precise overview of language-specific percentages of legal text in mC4 in Figure \ref{fig:percentage_legal}.

\begin{figure}[ht!]
    \centering
    \includegraphics[width=\columnwidth]{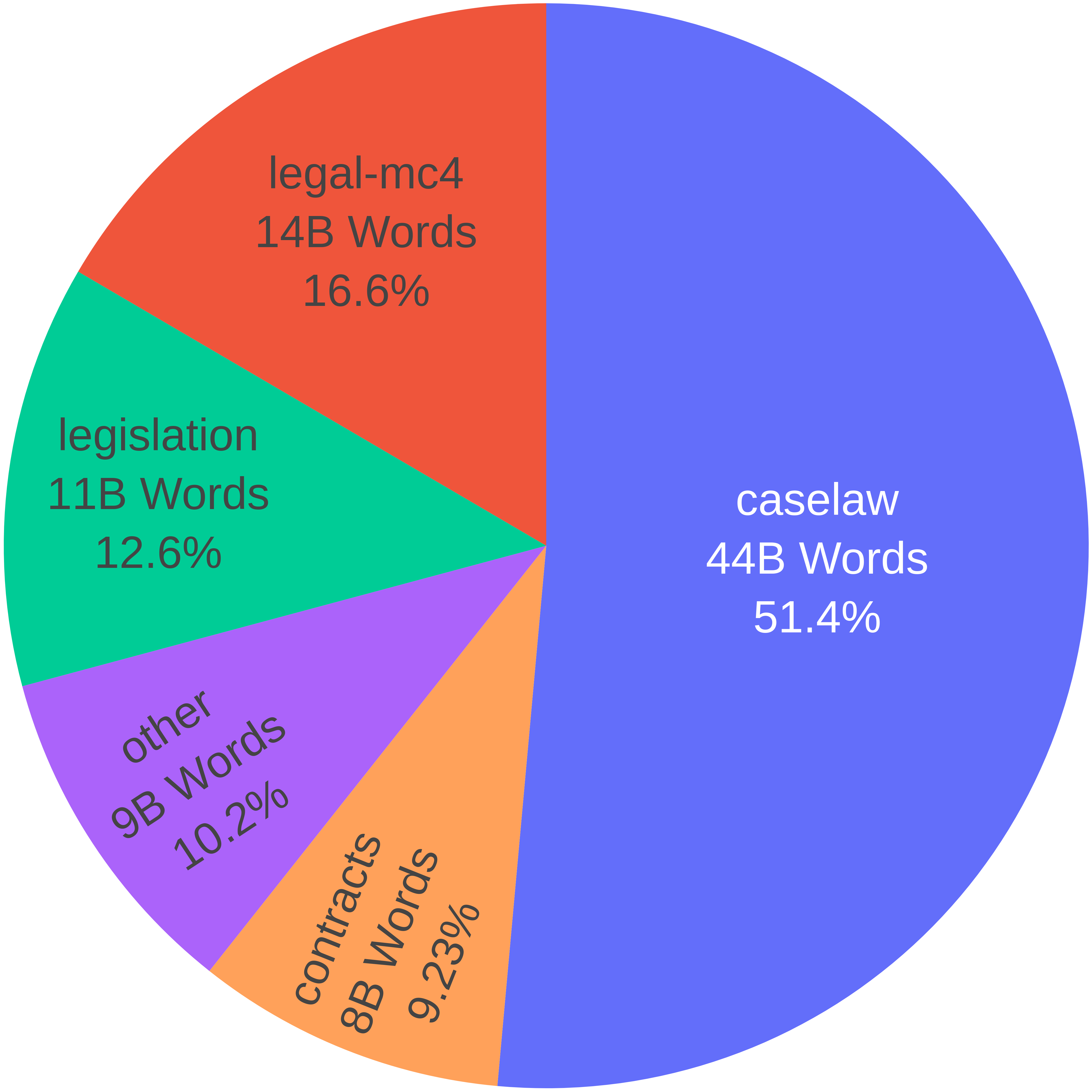}
    \caption{\textsc{MultiLegalPile} Text Type Distribution}
    \label{fig:text_type_distribution}
    \vspace{-5mm}
\end{figure}

\paragraph{Compiling Native \textsc{MultiLegalPile}}
To compile the corpus, we scraped several sources containing legal language materials. Our search was conducted in a loose manner, meaning that when we found a suitable source with legal text data, we included it in our corpus. It is important to note that we do not claim completeness, as we were unable to perform quality analysis for all available languages. For a detailed overview of sources used for the Native \textsc{MultiLegalPile} corpus, please refer to Table \ref{tab:multi_legal_pile}.
Most sources offered direct data download links. For inconsistently formatted data, we converted them to a unified format like jsonl. The post-processing steps involved performing various tasks depending on the initial data format. For example, in the case of CASS\footnote{\url{https://echanges.dila.gouv.fr/OPENDATA/CASS}}, we extracted the textual data from XML tags. 


\paragraph{Curating Eurlex Resources}


To curate the Eurlex resources, we utilized the eurlex R package \cite{eurlex_R} to generate SPARQL queries and download the data. Subsequently, we converted the data into a format more amenable to handling large datasets using Python. 

\paragraph{Integrating Pile of Law}
\citet{henderson_pile_2022} released a large corpus of diverse legal text in English mainly originating from the US. We integrated the latest version with additional data (from January 8, 2023) into our corpus.

\begin{figure*}[ht]
    \centering
    \includegraphics[width=\textwidth]{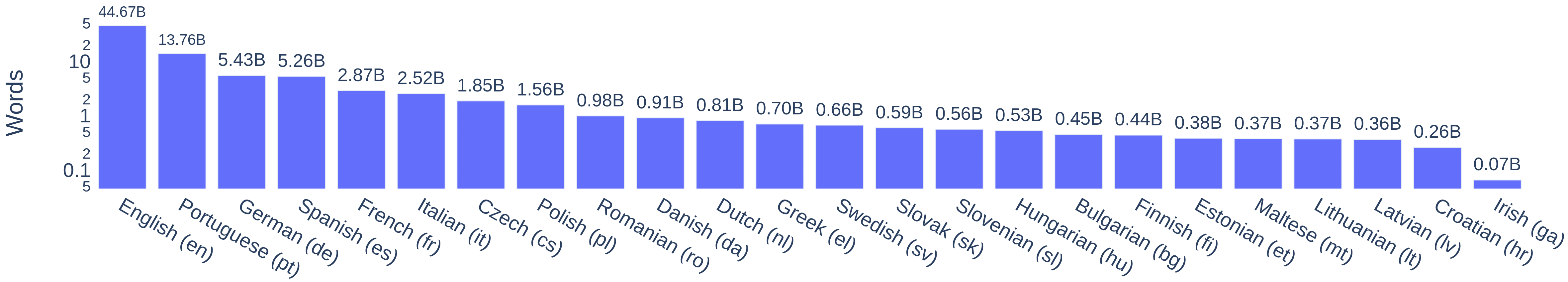}
    \vspace{-6mm}
    \caption{\textsc{MultiLegalPile} Language Distribution (Note the log-scaled y-axis)}
    \label{fig:language_distribution}
\end{figure*}

\subsection{Description}
\label{sec:description}

\textsc{MultiLegalPile} consists of four subsets: a) Native Multi Legal Pile (112 GB), b) Eurlex Resources 
(179 GB), c) Legal MC4 
(106 GB) and d) Pile of Law 
\cite{henderson_pile_2022} (292 GB).
Figure~\ref{fig:language_distribution} details the distribution of languages. Due to Pile of Law integration, English dominates, comprising over half the words.
Figure~\ref{fig:text_type_distribution} shows text type distribution. Caselaw comprises over half the corpus, due to the good public access to court rulings especially in common law countries. Even in civil law countries, where legislation is crucial, caselaw often outnumbers legislation, as seen in the Swiss case in Table~\ref{tab:multi_legal_pile}. Publicly available contracts are scarce, contributing less than 10\% to the corpus, despite potentially making up most existing legal texts (from the private sector). Note that most contracts in our corpus originate from the US or EU international treaties.  
Table \ref{tab:multi_legal_pile} in Appendix~\ref{sec:dataset_details} provides additional information on \textsc{MultiLegalPile}, including sources and licenses.

\subsection{Licenses and Usage of \textsc{MultiLegalPile}}
The Creative Commons Attribution-NonCommercial-ShareAlike 4.0 International (CC BY-NC-SA 4.0) license applied to the \textsc{MultiLegalPile} corpus depends on the upstream licenses of the data subsets described above. 

First, our \textit{Native Multi Legal Pile} consists of data sources with different licenses. They range from restrictive licenses such as CC BY-NC-SA 4.0 up to the most liberal Creative Commons Zero (CC0) license, which, in essence, releases the data into the public domain. Many sources, however, do not explicitly state the license used for the available data. We assume that such data sources allow pretraining usage, since the creators are usually public agencies such as courts and administrations. Such legislation and caselaw is usually not protected by copyright law. Table~\ref{tab:multi_legal_pile} provides an overview of the license or copyright situation for each of the 29 sources in the Native Multi Legal Pile.
Second, the \textit{Eurlex Resources} is CC BY 4.0 licensed by the European Union\footnote{\url{https://eur-lex.europa.eu/content/legal-notice/legal-notice.html}}, thus posing no legal issues for pretraining.
Third, the \textit{Legal mC4} corpus was created by filtering multilingual C4 \cite{xue_mt5_2021} for legal content as described above. As mC4\footnote{\url{https://huggingface.co/datasets/mc4}} is licensed under ODC-BY, we also release the filtered Legal mC4 corpus under the same license.
Finally, the \textit{Pile of Law} \cite{henderson_pile_2022} is published under CC BY-NC-SA 4.0 and the dataset is unaltered, thus preserving the license.

Usage of the \textsc{MultiLegalPile} corpus is presumably possible for pretraining of NLP models. In general, we assume that the fair use doctrine allows employing the data for legal NLP models because the results are rather transformative \cite{henderson2023foundation}. Nevertheless, copyright issues in generative AI remain an unresolved problem for the moment. Several court cases are currently pending, such as Getty Images suing Stability AI for intellectual property infringement \cite{sag2023copyright}.

\section{Pretraining Legal Models}

\begin{table*}
  \centering
  \footnotesize
  \resizebox{\textwidth}{!}{
  \begin{tabular}{llrrrrr}
    \toprule
    \textbf{Model} & \textbf{Source} & \textbf{Params} & \textbf{Vocab} & \textbf{Specs}  & 
  \textbf{Corpus} & \textbf{\# Langs} \\
    \midrule
    MiniLM              & \citet{wang_minilm_2020}                  & 118M & 250K & 1M steps / BS 256       & 2.5TB CC100 & 100\\
    DistilBERT          & \citet{sanh_distilbert_2020}              & 135M & 120K & BS up to 4000           & Wikipedia   & 104\\
    mDeBERTa-v3         & \citet{he_deberta_2021, he_debertav3_2021}& 278M & 128K & 500K steps / BS 8192    & 2.5TB CC100 & 100\\
    XLM-R base          & \citet{conneau-etal-2020-unsupervised}         & 278M & 250K & 1.5M steps / BS 8192    & 2.5TB CC100 & 100\\
    XLM-R large         & \citet{conneau-etal-2020-unsupervised}         & 560M & 250K & 1.5M steps / BS 8192    & 2.5TB CC100 & 100\\
    \midrule
    Legal-XLM-R-base    & ours                                      & 184M & 128K & 1M steps / BS 512     & 689GB MLP   & 24\\
    Legal-XLM-R-large   & ours                                      & 435M & 128K & 500K steps / BS 512     & 689GB MLP   & 24\\
    Legal-XLM-LF-base   & ours                                      & 208M & 128K & 50K steps / BS 512      & 689GB MLP   & 24\\
    Legal-mono-R-base   & ours                                      & 111M & 32K  & 200K steps / BS 512     & 689GB MLP   & 1\\
    Legal-mono-R-large   & ours                                     & 337M & 32K  & 500K steps / BS 512     & 689GB MLP   & 1\\
    \bottomrule
  \end{tabular}
  }
  \caption{Models: All models process up to 512 tokens, except Legal-XLM-LF-base (4096 tokens). BS is short for batch size.  MLP is short for \textsc{MultiLegalPile.} Params is the total parameter count (including embedding layer).}
  \label{tab:models}
  \vspace{-5mm}
\end{table*}

As part of this study, we release 2 new multi-lingual legal-oriented PLMs, dubbed Legal-XLM-Rs, trained on the newly introduced \textsc{MultiLegalPile} corpus (Section~\ref{sec:multilegalpile}). For the newly released Legal-XLM-Rs we followed a series of best-practices in the \ac{LM} development literature:\vspace{1mm}

\noindent (a) We warm-start (initialize) our models from the original XLM-R checkpoints (base or large) of \citet{conneau_cross-lingual_2019}. Model recycling is a standard process followed by many~\cite{wei-etal-2022, instructgpt} to benefit from starting from an available ``well-trained'' PLM, rather from scratch (random). XLM-R was trained on 2.5TB of cleaned CommonCrawl data in 100 languages. 

\noindent (b) We train a new tokenizer of 128K BPEs on the training subsets of \textsc{MultiLegalPile} to better cover legal language across all available legal systems and languages. However, we reuse the original XLM-R embeddings for all lexically overlapping tokens \cite{pfeiffer-etal-2021-unks}, i.e., we warm-start word embeddings for tokens that already exist in the original XLM-R vocabulary, and use random ones for the rest. 
Similarly to \citet{liang_xlm-v_2023} who increased the vocabulary size from around 2.5K tokens per language (250K for 100 languages) to around 10K (1M for 100 languages), we increased to around 5K (128K for 24 languages), thus roughly doubling compared to XLM-R.

\noindent (c) We continue pretraining our models on the diverse \textsc{MultiLegalPile} corpus with batches of 512 samples for an additional 1M/500K steps for the base/large model. We do initial warm-up steps for the first 5\% of the total training steps with a linearly increasing learning rate up to $1e\!-\!4$, and then follow a cosine decay scheduling, following recent trends. For half of the warm-up phase (2.5\%), the Transformer encoder is frozen, and only the embeddings, shared between input and output (MLM), are updated. We also use an increased 20/30\% masking rate for base/large models respectively, where 100\% of token predictions are based on masked tokens, compared to \citet{devlin-etal-2019-bert}\footnote{\citeauthor{devlin-etal-2019-bert} -- and many other follow-up work -- used a 15\% masking ratio, and a recipe of 80/10/10\% of predictions made across masked/randomly-replaced/original tokens.}, based on the findings of~\citet{wettig2022should}.

\noindent (d) For both training the tokenizer and our legal models, we use a sentence sampler with exponential smoothing of the sub-corpora sampling rate following \citet{conneau_cross-lingual_2019} and \citet{raffel_exploring_2020}, since there is a disparate proportion of tokens across sub-corpora and languages (Figures~\ref{fig:source_distribution} and~\ref{fig:language_distribution}) and we aim to preserve per-corpus and language capacity, i.e., avoid overfitting to the majority (approx. 50\% of the total number of tokens) US-origin English texts.

\noindent (e) We consider mixed cased models, i.e., both upper- and lowercase letters covered, similar to recently developed large PLMs~\cite{conneau_cross-lingual_2019, raffel_exploring_2020, brown-etal-gpt3}.

To better account for long contexts often found in legal documents, we continue training the base-size multilingual model on long contexts (4096 tokens) with windowed attention (128 tokens window size) \cite{beltagy_longformer_2020} for 50K steps, dubbing it Legal-XLM-LF-base. We use the standard 15\% masking probability and increase the learning rate to $3e\!-\!5$ before decaying but otherwise use the same settings as for training the small-context models. 

In addition to the multilingual models, we also train 24 monolingual models on each of the language-specific subsets of the corpus. Except for choosing a smaller vocab size of 32K tokens, we use the same settings as for the multilingual models. Due to resource constraints, we only train base-size models and stop training at 200K steps. Due to limited data available in some low-resource languages, these models sometimes do multiple passes over the data. Because of plenty of data and to achieve a better comparison on LexGLUE, we continued training the English model for 1M steps and also trained a large-size model for 500K steps. See Table~\ref{tab:model_details} in appendix~\ref{sec:training_details} for an overview.

We make all our models publicly available alongside all intermediate checkpoints (every 50K/10K 
training steps for RoBERTa/Longformer 
models) on the Hugging Face Hub.\footnote{
Link will be released upon acceptance.
}

\section{Evaluating on Legal Benchmarks}

In the absence of established legal benchmarks for generative tasks and our focus on pretraining encoder-only models, we select two established legal benchmarks involving challenging text classification and named entity recognition tasks involving long documents: LEXTREME and LexGLUE.

\begin{table*}[!ht]
\centering
\resizebox{\textwidth}{!}{
\begin{tabular}{lrrrrrrrrrrrr}
\toprule
       \bf{Model} &  \bf{BCD} &  \bf{GAM} &  \bf{GLC} &  \bf{SJP} &  \bf{OTS} &  \bf{C19} &  \bf{MEU} &  \bf{GLN} &  \bf{LNR} &  \bf{LNB} &  \bf{MAP} & \bf{Agg.} \\
\midrule
           MiniLM &      53.0 &      73.3 &      42.1 &      67.7 &      44.1 &       5.0 &      29.7 &      74.0 &      84.5 &      93.6 &      57.8 &      56.8 \\
       DistilBERT &      54.5 &      69.5 &      62.8 &      66.8 &      56.1 &      25.9 &      36.4 &      71.0 &      85.3 &      89.6 &      60.8 &      61.7 \\
      mDeBERTa-v3 &      60.2 &      71.3 &      52.2 &      69.1 &      66.5 &      29.7 &      37.4 &      73.3 &      85.1 &      94.8 &      67.2 &      64.3 \\
       XLM-R-base &      63.5 &      72.0 &      57.4 &      69.3 &      67.8 &      26.4 &      33.3 & \bf{74.6} & \bf{85.8} &      94.1 &      62.0 &      64.2 \\
      XLM-R-large &      58.7 &      73.1 &      57.4 &      69.0 & \bf{75.0} &      29.0 & \bf{42.2} &      74.1 &      85.0 & \bf{95.3} &      68.0 &      66.1 \\
      \midrule
 Legal-XLM-R-base &      62.5 &      72.4 &      68.9 &      70.2 &      70.8 &      30.7 &      38.6 &      73.6 &      84.1 &      94.1 & \bf{69.2} &      66.8 \\
Legal-XLM-R-large &      63.3 &      73.9 &      59.3 &      70.1 &      74.9 & \bf{34.6} &      39.7 &      73.1 &      83.9 &      94.6 &      67.3 &      66.8 \\
Legal-XLM-LF-base & \bf{72.4} & \bf{74.6} & \bf{70.2} & \bf{72.9} &      69.8 &      26.3 &      33.1 &      72.1 &      84.7 &      93.3 &      66.2 & \bf{66.9} \\
\bottomrule
\end{tabular}
}
\caption{Dataset aggregate scores (macro-F1) for multilingual models on LEXTREME with the best scores in \textbf{bold}.}
\label{tab:lextreme_dataset_agg}
\end{table*}

{
\setlength{\tabcolsep}{3pt}
\begin{table*}[!ht]
\fontsize{7pt}{7pt}\selectfont
\centering
\resizebox{\textwidth}{!}{
\begin{tabular}{lrrrrrrrrrrrrrrrrrrrrrrrrr}
\toprule
                    \bf{Model} &   \bf{bg} &   \bf{cs} &   \bf{da} &   \bf{de} &   \bf{el} &   \bf{en} &   \bf{es} &   \bf{et} &   \bf{fi} &   \bf{fr} &   \bf{ga} &   \bf{hr} &   \bf{hu} &   \bf{it} &   \bf{lt} &   \bf{lv} &   \bf{mt} &   \bf{nl} &   \bf{pl} &   \bf{pt} &   \bf{ro} &   \bf{sk} &   \bf{sl} &   \bf{sv} & \bf{Agg.} \\
\midrule
MiniLM &      52.7 &      48.6 &      42.8 &      54.6 &      50.3 &      34.3 &      40.1 &      46.3 &      42.2 &      39.0 &      42.8 &      29.7 &      29.6 &      40.5 &      44.2 &      40.8 &      40.8 &      29.5 &      22.7 &      61.6 &      59.6 &      44.3 &      30.0 &      43.4 &      40.5 \\
       DistilBERT &      54.2 &      48.6 &      46.0 &      60.1 &      58.8 &      48.0 &      50.0 &      48.8 &      49.6 &      47.9 &      51.4 &      35.9 &      31.2 &      50.1 &      51.9 &      41.5 &      44.4 &      34.6 &      34.5 &      63.2 &      63.8 &      51.3 &      36.2 &      50.1 &      46.7 \\
      mDeBERTa-v3 &      54.1 &      51.3 &      51.7 &      63.6 &      57.7 &      50.7 &      53.3 &      50.8 &      54.6 &      49.2 &      54.9 &      37.4 &      37.5 &      55.1 &      53.9 &      47.0 &      52.5 & 42.1 &      41.0 &      65.7 &      65.3 &      55.4 &      37.5 &      56.1 &      50.5 \\
       XLM-R-base &      56.4 &      48.3 &      48.3 &      60.6 &      57.6 &      50.1 &      47.2 &      46.7 &      48.6 &      49.4 &      50.1 &      33.6 &      32.8 &      53.4 &      50.0 &      44.1 &      43.8 &      35.2 &      41.3 &      66.1 &      63.7 &      45.3 &      33.7 &      50.0 &      47.1 \\
      XLM-R-large &      \bf{59.9} &      56.0 &      \bf{56.3} &      65.4 &      60.8 &      56.2 &      \bf{56.6} &      56.5 &      \bf{56.9} &      51.4 &      55.4 & 42.5 &      38.1 &      \bf{58.5} &      58.1 &      49.9 &      53.9 &      39.5 &      \bf{46.4} &      \bf{68.6} &      \bf{66.8} &      \bf{57.9} & 42.4 &      \bf{59.1} &      \bf{53.7} \\

\midrule
Legal-XLM-R-base &      55.6 & \bf{58.8} &      50.4 &      63.6 & \bf{63.7} &      66.8 &      56.3 & \bf{57.0} &      52.6 &      50.1 &      56.6 &      38.7 & \bf{56.5} &      56.1 &      57.2 &      49.1 &      56.0 &      41.6 &      43.9 &      68.2 &      66.1 &      55.6 &      38.6 &      54.9 &      53.5 \\
Legal-XLM-R-large &      57.8 &      55.6 &      50.4 & \bf{65.7} &      60.7 & \bf{69.3} &      55.7 &      54.5 &      56.6 & \bf{53.3} & \bf{57.2} &      39.7 &      39.1 &      58.1 & \bf{60.6} &      48.4 &      57.2 &      39.4 &      45.5 &      67.3 &      65.5 &      49.3 &      39.7 &      56.4 &      53.6 \\
Legal-XLM-LF-base &      54.4 &      49.3 &      48.1 &      64.0 &      60.5 &      52.8 &      49.2 &      52.2 &      48.2 &      48.5 &      55.4 &      33.0 &      34.7 &      54.6 &      54.8 &      45.2 &      52.5 &      40.1 &      40.6 &      68.3 &      64.1 &      48.4 &      33.0 &      51.3 &      48.9 \\
\midrule
NativeLegalBERT &           - &           - &           - &           - &           - &        53.1 &        46.9 &           - &           - &           - &           - &           - &           - &        45.3 &           - &           - &           - &           - &           - &           - &        59.0 &           - &           - &           - &          51.1 \\
NativeBERT &        54.8 &    57.3 &        51.2 &        63.0 &     62.3 &        52.0 &        42.6 &        47.2 &        52.4 &        49.4 &        50.1 &           - &        37.4 &        47.1 &           - &           - &           - &        37.0 &        40.5 &        66.5 &        63.1 &        44.8 &           - &        55.1 &          50.2 \\
Legal-mono-R-base &           55.9 &           49.5 &           51.5 &           61.3 &           61.3 &        50.5 &        52.1 &           53.5 &           53.6 &           51.1 &           52.2 &           \bf{44.1} &           54.1 &        51.8 &           55.5 &           \bf{50.0} &           \bf{59.1} &           \bf{54.3} &           34.4 &           67.1 &        61.5 &           48.8 &           \bf{53.4} &           58 &          53.5 \\
\bottomrule
\end{tabular}
}
\caption{Language aggregate scores (macro-F1) for multilingual models on LEXTREME with the best scores in \textbf{bold}. For each language, we list the top-performing monolingual legal and non-legal models under \textit{NativeLegalBERT} and \textit{NativeBERT}, and our legal models under \textit{Legal-mono-R-base}. Missing values signify no suitable models found.}
\label{tab:lextreme_language_agg}
\end{table*}
}

\subsection{Benchmark Description}
Below, we briefly describe each dataset and refer the reader to the original works for more details. 

\paragraph{LEXTREME} \cite{niklaus_lextreme_2023} is a multilingual legal benchmark. It includes five single label text classification datasets,
three multi label text classification datasets 
and four \ac{NER} datasets.

\textbf{Brazilian Court Decisions (BCD)} \cite{lage-freitas_predicting_2022} is from the State Supreme Court of Alagoas (Brazil) and involves predicting case outcomes and judges' unanimity on decisions.
\textbf{German Argument Mining (GAM)} \cite{urchs_design_2021} contains 200 German court decisions for classifying sentences according to their argumentative function.
\textbf{Greek Legal Code (GLC)} \cite{Papa2021m} tackles topic classification of Greek legislation documents. Tasks involve predicting topic categories at volume, chapter, and subject levels.
\textbf{Swiss Judgment Prediction (SJP)} \cite{niklaus_swiss-judgment-prediction_2021} focuses on predicting the judgment outcome from 85K cases from the Swiss Federal Supreme Court.
\textbf{Online Terms of Service (OTS)} \cite{drawzeski_corpus_2021} contains 100 contracts for detecting unfair clauses with the tasks of classifying sentence unfairness levels and identifying clause topics.
\textbf{COVID19 Emergency Event (C19)} \cite{tzia2021}: consists of legal documents from several European countries related to COVID-19 measures where models identify the type of measure described in a sentence.
\textbf{MultiEURLEX (MEU)} \cite{Chalkidis2021MultiEURLEXA} is a corpus of 65K EU laws annotated with EUROVOC taxonomy labels. Task involves identifying labels for each document.
\textbf{Greek Legal NER (GLN)} \cite{angelidis_named_2018} is a dataset for \ac{NER} in Greek legal documents.
\textbf{LegalNERo (LNR)} \cite{pais_vasile_2021_4922385} tackles \ac{NER} in Romanian legal documents.
\textbf{LeNER BR (LNB)} \cite{luz2018a} addresses \ac{NER} in Brazilian legal documents. 
\textbf{MAPA (MAP)} \cite{baisa-etal-2016-european} is a multilingual corpus based on EUR-Lex for \ac{NER} annotated at a coarse-grained and fine-grained level.

\paragraph{LexGLUE} \cite{chalkidis_lexglue_2022} is a legal benchmark covering two single-label and,
four multi-label text classification datasets, 
and a multiple choice question answering dataset. 

\textbf{ECtHR Tasks A \& B} \cite{chalkidis-etal-2019-neural, chalkidis-etal-2021-paragraph} contain approx. 11K cases from the European Court of Human Rights (ECtHR) public database. Based on case facts, Task A predicts violated articles and Task B allegedly violated articles of the European Convention of Human Rights (ECHR).
\textbf{SCOTUS} \cite{Spaeth2020} combines information from US Supreme Court (SCOTUS) opinions with the Supreme Court DataBase (SCDB). The task is to classify court opinions into 14 issue areas. 
\textbf{EUR-LEX} \cite{Chalkidis2021MultiEURLEXA} contains 65K EU laws from the EUR-Lex portal, annotated with EuroVoc concepts. The task is to predict EuroVoc labels for a given document. 
\textbf{LEDGAR} \cite{tuggener-etal-2020-ledgar} contains approx. 850K contract provisions from the US Securities and Exchange Commission (SEC) filings. The task is to classify contract provisions into categories. 
\textbf{UNFAIR-ToS} \cite{Lippi2019} contains 50 Terms of Service (ToS) from online platforms, annotated with types of unfair contractual terms. The task is to predict unfair types for a given sentence. 
\textbf{CaseHOLD} \cite{10.1145/3462757.3466088} contains approx. 53K multiple choice questions about holdings of US court cases. The task is to identify the correct holding statement out of five choices.

\begin{table*}[!ht]
\centering
\resizebox{\textwidth}{!}{
\begin{tabular}{lrrrrrrrr}
\toprule
                      \bf{Model} & \bf{ECtHR-A} & \bf{ECtHR-B} & \bf{SCOTUS} & \bf{EUR-LEX} & \bf{LEDGAR} & \bf{UNFAIR-ToS } & \bf{CaseHOLD} & \bf{Agg.} \\
\midrule
    TFIDF+SVM * &         48.9 &         63.8 &        64.4 &         47.9 &        81.4 &             75.0 &          22.4 &      49.0 \\
         BERT * &         63.6 &         73.4 &        58.3 &         57.2 &        81.8 &             81.3 &          70.8 &      68.2 \\
      DeBERTa * &         60.8 &         71.0 &        62.7 &         57.4 &        83.1 &             80.3 &          72.6 &      68.5 \\
 RoBERTa-base * &         59.0 &         68.9 &        62.0 &         57.9 &        82.3 &             79.2 &          71.4 &      67.5 \\
RoBERTa-large * &         67.6 &         71.6 &        66.3 &         58.1 &   \bf{83.6} &             81.6 &          74.4 &      70.9 \\
\midrule
   Longformer * &         64.7 &         71.7 &        64.0 &         57.7 &        83.0 &             80.9 &          71.9 &      69.5 \\
      BigBird * &         62.9 &         70.9 &        62.0 &         56.8 &        82.6 &             81.3 &          70.8 &      68.4 \\
      \midrule
   Legal-BERT * &         64.0 &         74.7 &        66.5 &         57.4 &        83.0 &        \bf{83.0} &          75.3 &      70.8 \\
 CaseLaw-BERT * &         62.9 &         70.3 &        65.9 &         56.6 &        83.0 &             82.3 &          75.4 &      69.7 \\
 \midrule
          Legal-en-R-base (ours) &         65.2 &         73.7 &        66.4 &    \bf{59.2} &        82.7 &             78.7 &          73.3 &      70.5 \\
         Legal-en-R-large (ours) &    \bf{70.3} &    \bf{77.0} &   \bf{67.7} &         58.4 &        82.5 &             82.4 &     \bf{77.0} & \bf{72.7} \\
         Legal-XLM-R-base (ours) &         64.8 &         73.9 &        63.9 &         58.2 &        82.8 &             79.6 &          71.7 &      69.7 \\
        Legal-XLM-R-large (ours) &         68.2 &         74.2 &        67.5 &         58.4 &        82.7 &             79.9 &          75.1 &      71.4 \\
        Legal-XLM-LF-base (ours) &         67.9 &         76.2 &        61.6 &         59.1 &        82.1 &             78.9 &          72.0 &      70.2 \\
\bottomrule
\end{tabular}
}
\caption{Results on LexGLUE (macro-F1) with the best scores in \textbf{bold}. Results marked with * are from \citet{chalkidis_lexglue_2022}. Similar to LEXTREME, we calculate the aggregate score as the harmonic mean of dataset results.}
\vspace{-5mm}
\label{tab:lex_glue_results}
\end{table*}

\subsection{Experimental Setup}

To ensure comparability, we followed the experimental setups described in the original papers \cite{niklaus_lextreme_2023, chalkidis_lexglue_2022} using hierarchical transformers for datasets where the sequence length of most documents exceeds the maximum sequence length of the model \cite{aletras_predicting_2016, niklaus_empirical_2022}. The hyperparameters used for running experiments on each dataset are provided in Table \ref{tab:hyperparameters} in the appendix. We follow previous work \cite{niklaus_lextreme_2023, chalkidis_lexglue_2022} and do not tune hyperparameters.

All scores are macro-F1 scores, equally weighing each class for fairness in unbalanced datasets. To obtain Table \ref{tab:lex_glue_results}, we follow \citet{chalkidis_lexglue_2022}, running five repetitions with different random seeds (1-5) and report test scores from the best-performing seed on the development data. For values in Tables \ref{tab:lextreme_dataset_agg} and \ref{tab:lextreme_language_agg}, we follow the procedure in \citet{niklaus_lextreme_2023}, taking the harmonic mean of the results of 3 random seeds (1-3).
We calculate the dataset aggregate in Table \ref{tab:lextreme_dataset_agg} by successively taking the harmonic mean of (i) the languages in the configurations (e.g., de,fr,it in SJP), (ii) configurations within datasets (e.g., OTS-UL, OTS-CT in OTS), and (iii) datasets in LEXTREME (BCD, GAM).
The language aggregate score in Table \ref{tab:lextreme_language_agg} is computed similarly: by taking the harmonic mean of (i) configurations within datasets, (ii) datasets for each language (e.g., MAP, MEU for lv), and (iii) languages in LEXTREME (bg,cs).
We show an overview of the models evaluated in Table~\ref{tab:models}.

Note that most \acp{LLM} are predominantly trained on English and Chinese with the exception of mT5 \cite{xue_mt5_2021} and BLOOM \cite{scao_bloom_2022} (more than 95\% of LLaMA's pretraining corpus is English \cite{touvron_llama_2023}). Because LEXTREME and LexGLUE consist of NLU tasks, we compare to encoder-only \acp{LM} only. 

\subsection{Evaluation on LEXTREME}
We evaluate our models on LEXTREME \cite{niklaus_lextreme_2023} and show results across datasets in Table~\ref{tab:lextreme_dataset_agg} and across languages in Table~\ref{tab:lextreme_language_agg}. 

We notice that our Legal-XLM-R-base model is on par with XLM-R large even though it only contains 33\% of the parameters (184M vs 560M). All our models outperform XLM-R large on the dataset aggregate score. Our base model sets a new SotA on MAPA (MAP), the large model on CoViD 19 emergency event (C19) and the Longformer on Brazilian court decisions (BCD), German argument mining (GAM), Greek legal code (GLC) and Swiss judgment prediction (SJP). 
Surprisingly, the legal models slightly underperform in three NER tasks (GLN, LNR, and LNB). Sensitivity to hyperparameter choice could be a reason for this underperformance (we used the same hyperparameters for all models without tuning due to limited compute resources). 
We see the largest improvements over prior art in BCD (72.4 vs. 63.5) and in GLC (70.2 vs 62.8). Maybe these tasks are particularly hard, and therefore legal in-domain pretraining helps more. For BCD especially, the large amount of Brazilian caselaw in the pretraining corpus may offer an additional explanation.

The monolingual models underperform their base model XLM-R base only in Italian, Polish, and Romanian. In some languages the monolingual model even outperforms XLM-R base clearly (Estonian, Croatian, Hungarian, Latvian, Maltese, Dutch, Slovenian, and Swedish), and in five of them even set the new SotA for the language, sometimes clearly outperforming all other models (the Dutch model even outperforms its closest competitor mDeBERTa-v2 by 11.2 macro F1 and its base model XLM-R by almost 20 macro F1). These languages are all in the lower end of the data availability in the \textsc{MultiLegalPile} with the richest language (Dutch) containing only 810M words (see Figure~\ref{fig:language_distribution}). Pretraining a monolingual model on in-domain data may therefore be worth it, especially in low-resource languages.




Even though our legal Longformer model performs best on the dataset level, it performs much worse on the language level, possibly due to its lower scores in the most multilingual tasks MEU, MAP and C19 (24, 24 and 6 languages, respectively). 
Our legal base and large models achieve SotA in some languages, and are in aggregate almost as robust across languages as XLM-R. 

Computing the final LEXTREME scores (harmonic mean of dataset aggregate and language aggregate scores), we find that the Legal-XLM-R-large is the new SotA on LEXTREME with a score of 59.5 vs 59.4 for Legal-XLM-R-base and 59.3 for XLM-R large. The legal Longformer's LEXTREME score (56.5) is not competitive due to its low language aggregate score.


\subsection{Evaluation on LexGLUE}

We evaluate our English and multilingual models on LexGLUE \cite{chalkidis_lexglue_2022} and compare with baselines (see Table~\ref{tab:lex_glue_results}). 
Our models excel on the ECtHR, SCOTUS, EUR-LEX, and CaseHOLD tasks, setting new SotA. In the other two tasks, our models match general-purpose models such as RoBERTa. A reason for slight underperformance of the legal models in the LEDGAR and especially the Unfair ToS tasks may be the relatively low availability of contracts in the \textsc{MultiLegalPile}.

\section{Conclusions and Future Work}

\paragraph{Conclusions}
Due to a general lack of multilingual pretraining data especially in specialized domains such as law, we curate a large-scale high-quality corpus in 24 languages from 17 jurisdictions. 
We continue pretraining XLM-R checkpoints on our data, achieving a new SotA for base and large models on the LEXTREME benchmark and vastly outperforming previous methods in Greek legal code. We turn our XLM-R base model into a Longformer and continue pretraining on long documents. It reaches a new SotA in four LEXTREME datasets and reaches the overall highest dataset aggregate score. 
Monolingual models achieve huge gains over their base model XLM-R in some languages and even set language specific SotA in five languages outperforming other models by as much as 11 macro F1. 
On LexGLUE our English models reach SotA in five out of seven tasks with the large model achieving the highest aggregate score. 
To conclude, following best practices in continued pretraining on our comprehensive multilingual legal corpus establishes new state-of-the-art across multiple datasets and languages, significantly enhancing performance in legal text analysis.

\paragraph{Future Work}
We focused on the 24 EU languages, but in the future, we would like to expand the corpus in terms of languages and jurisdictions covered. Especially in China there exist many accessible sources suitable to extend the corpus. Additionally, we would like to find out whether our findings on in-domain pretraining hold for multi-billion generative models. Finally, a detailed examination of the contents of the \textsc{MultiLegalPile} could provide valuable insights into its structure and efficacy in enhancing legal language models.

\section*{Ethics Statement}
This study focuses on evaluating legal-specific LMs from multiple aspects, expanding the dialogue to aid in creating support technologies for both legal professionals and the general public. This area represents a vital field for research, as emphasized by \citet{tsarapatsanis_ethical_2021}, aiming to enhance legal services and make legal knowledge more accessible. The study also aims to shed light on the multifaceted limitations that need addressing to ensure the responsible and ethical application of legal-oriented technologies.

In pursuit of these goals, we introduce novel resources encompassing a range of legal systems. These resources are designed to construct new models that more accurately reflect legal nuances and evaluate their effectiveness more precisely. All resources created and shared in this work are derived from data that is publicly accessible, often distributed across various online platforms.

\section*{Limitations}

We did not perform deduplication, thus data from legal mC4 might be present in other parts. However, \citet{muennighoff_scaling_2023} suggest that data duplication does not degrade performance during pretraining for up to four epochs. Overlap between other subsets is highly unlikely, since they originate from completely different jurisdictions.


Due to limited compute, we were not able to pretrain a large generative model and leave this to future work.


\section{Bibliographical References}\label{sec:reference}

\bibliography{anthology, custom, references}
\bibliographystyle{lrec-coling2024-natbib}

\newpage

\section{Use of AI assistants}
We used ChatGPT and Grammarly for improving the grammar and style of our writing.

\section{Additional Related Work}
\label{sec:additional_related_work}

\subsection{General Pretraining Corpora}
The use of \acp{PLM} has become increasingly popular in NLP tasks, particularly with the advent of models such as BERT \cite{devlin-etal-2019-bert} that can be finetuned for specific applications. One key factor in the success of pretraining is the availability of large and diverse text corpora, which can help the model learn the nuances of natural language. In the following, we discuss large-scale general-purpose text corpora used for pretraining.

One of the earliest widely-used datasets is the One Billion Word Language Model Benchmark (LM1B) \cite{chelba2014billion}. It was created by extracting one billion words from web pages to evaluate novel language modeling techniques. It has been used, among others, to evaluate GPT-2 \cite{radford_language_2019}.

Wikipedia is a commonly used multilingual dataset for pretraining language models, and has been used to pretrain BERT \cite{devlin-etal-2019-bert}, MegatronBERT \cite{shoeybi2020megatronlm}, T5 \cite{raffel_exploring_2020}, and GPT-3 \cite{brown-etal-gpt3}, among others. 

Based on Wikipedia, \citet{merity2016pointer} created WikiText by selecting articles fitting the Good or Featured article criteria. The dataset contains 103M words and has two versions: WikiText2 and the larger WikiText103. It has been used to pretrain models like MegatronBERT \cite{shoeybi2020megatronlm} and GPT-2 \cite{radford_language_2019}.

The BookCorpus \cite{BookCorpus2015}, also known as the Toronto Books Corpus, is an English dataset used for pretraining BERT \cite{devlin-etal-2019-bert}, RoBERTa \cite{liu_roberta_2019}, and T5 \cite{raffel_exploring_2020}. It consists of almost 1B words from over 11K books collected from the web.

The Common Crawl corpus is a publicly available multilingual dataset of scraped web pages, regularly updated with new "snapshots". It has been used to pretrain GPT-3 \cite{brown-etal-gpt3} as well as XLM-R \cite{conneau-etal-2020-unsupervised}. One significant drawback of Common Crawl is the presence of uncleaned data, which includes a considerable amount of \say{gibberish or boiler-plate text like menus, error messages, or duplicate text} \cite{raffel_exploring_2020}. As a result, utilizing the Common Crawl dataset necessitates additional post-filtering and cleaning procedures. To address this issue, Raffel et al. \cite{raffel_exploring_2020} performed several cleaning steps on the April 2019 snapshot of Common Crawl, resulting in the creation of the Colossal Clean Crawled Corpus (C4), comprising 750 GB of English-language text. It was used for pretraining models such as T5 \cite{raffel_exploring_2020} and Switch Transformer \cite{Fedus2022}.

OpenWebText \cite{Gokaslan2019OpenWeb} openly replicates OpenAI's closed English WebText dataset \cite{radford_language_2019}, used to pretrain GPT-2 \cite{radford_language_2019}. WebText comprises over 8M documents with a combined text size of 40 GB. To ensure data uniqueness, any documents sourced from Wikipedia were excluded from WebText, as they are commonly utilized in other datasets. OpenWebText, on the other hand, consists of 38 GB of text data from 8M documents and was used for pretraining RoBERTa \cite{liu_roberta_2019} and MegatronBERT \cite{shoeybi2020megatronlm}.

News articles are also a common source for pretraining corpora. The RealNews dataset \cite{RealNews2019} is a large corpus extracted from Common Crawl, containing news articles from December 2016 to March 2019 (training) and April 2019 (evaluation), totaling 120 GB. It was used for pretraining MegatronBERT \cite{shoeybi2020megatronlm}. For pretraining RoBERTa, \citet{liu_roberta_2019} used an English subset of RealNews\footnote{\href{https://commoncrawl.org/2016/10/news-dataset-available}{https://commoncrawl.org/2016/10/news-dataset-available}}, comprising 63M English news articles crawled from September 2016 to February 2019.

The rise of \acp{LLM} brought about the creation of ever larger training datasets. The Pile \cite{ThePile2020} combines 22 distinct, well-curated datasets, such as Wikipedia (English), OpenWebText2 \cite{Gokaslan2019OpenWeb}, OpenSubtitles \cite{tiedemann-2016-finding} etc., encompassing 825 GB of data. Besides general-purpose textual datasets, it also contains domain-specific datasets, such as ArXiv (Science), FreeLaw (Legal), PubMed Abstracts (Biomedicine), and GitHub data (to improve code-related task performance \cite{ThePile2020}). GPT-2 \cite{radford_language_2019} and GPT-3 \cite{brown-etal-gpt3} were evaluated on this dataset.

\citet{touvron_llama_2023} compiled a substantial dataset from various publicly available sources, including CommonCrawl, C4, Github, Wikipedia, etc., totaling 1.4T tokens.  They trained the 13B-parameter LLaMA model using this dataset, surpassing the performance of the 175B-parameter GPT-3 on most benchmark tasks. However, the dataset itself is not publicly available. To address this, a collaborative effort resulted in the creation of the RedPajama-Data-1T\footnote{\url{https://github.com/togethercomputer/RedPajama-Data}} dataset, replicating LLaMA's dataset with a similar size of 1.2T tokens.

Some of the afore-mentioned datasets, such as Common Crawl, are used to pretrain multilingual versions of BERT, DistilBERT, RoBERTa etc. These models were pretrained on datasets that cover approximately 100 languages, thereby neglecting low-resource languages. \citet{imanigooghari2023glot500} addressed this by compiling Glot500, a 700 GB dataset covering 500 diverse languages, with a focus on low-resource ones. The Glot500-m model, pretrained on this dataset, outperformed the XLM-RoBERTa base model on six out of seven tasks.

\subsection{Domain Specific Corpora}

While pretraining on general-purpose text like Wikipedia and news articles shows promise, evidence suggests that pretraining on domain-specific text can enhance language model performance on related tasks \cite{Beltagy2019SciBERTAP, Gu2021, chalkidis-etal-2020-legal, niklaus_budgetlongformer_2022}. Domain-specific text corpora include texts specific to fields like medicine, law, or science.

Several studies have examined pretraining on scientific text corpora. \citet{Beltagy2019SciBERTAP} pretrained SciBERT, a BERT-based model, on a random subset of 1.14M papers sourced from Semantic Scholar. This collection comprises 18\% of computer science papers and 82\% of papers from the broader biomedical field. Similarly, PubMed and PubMedCentral are common sources for biomedical datasets. \citet{Gu2021} trained PubMedBERT using PubMed abstracts and PubMedCentral articles; BioBERT \cite{Lee2020} was pretrained similarly.  \citet{Johnson2016} compiled the Medical Information Mart for Intensive Care III (MIMIC-III) dataset, a large single-center database of critical care patients. 
\say{a large, single-center database comprising information relating to patients admitted to critical care units at a large tertiary care hospital}. 
\citet{Huang2019} used over 2 million de-identified clinical notes from this dataset to pretrain ClinicalBERT. These models outperformed general-purpose models on biomedical NLP tasks.

In the legal domain, similar strategies are observed. \citet{chalkidis-etal-2020-legal} collected 12 GB of diverse English legal texts, including legislation, court cases, and contracts. They pretrained LegalBERT on this dataset, showing SotA performance, especially in tasks requiring domain knowledge. Another study by \citet{10.1145/3462757.3466088} used the entire English Harvard Law case corpus (1965-2021) comprising 37 GB of text to pretrain CaseLaw-BERT.

Recently, \citet{chalkidis2023lexfiles} released LexFiles, an English legal corpus with 11 sub-corpora covering legislation and case law from six English-speaking legal systems (EU, Council of Europe, Canada, US, UK, India). The corpus contains approx.~6M documents or approx.~19B tokens. They trained two new legal English PLMs, showing improved performance in legal probing and classification tasks.

Efforts to pretrain legal language models also exist for Italian \cite{licari_italian-legal-bert_2022}, Romanian \cite{masala-etal-2021-jurbert}, and Spanish \cite{gutierrez-fandino_spanish_2021}. However, English dominates, underscoring the importance of compiling multilingual legal corpora.

\section{Training Details}
\label{sec:training_details}

For finetuning the pretrained models on the evaluation benchmarks we used the following NVIDIA GPUs: 24GB RTX3090, 32GB V100 and 80GB A100. We used v3-8 TPUs for pretraining. All our experiments were run on Linux machines (Debian).

\begin{table}
\centering
\resizebox{\columnwidth}{!}{
\begin{tabular}{lrr}
\toprule
Model Name & \# Steps & Vocab Size \\
\midrule
Legal-bg-R-base & 200K & 32K \\
Legal-hr-R-base & 200K & 32K \\
Legal-cs-R-base & 200K & 32K \\
Legal-da-R-base & 200K & 32K \\
Legal-nl-R-base & 200K & 32K \\
Legal-en-R-base & 200K & 32K \\
Legal-en-R-large & 500K & 32K \\
Legal-et-R-base & 200K & 32K \\
Legal-fi-R-base & 200K & 32K \\
Legal-fr-R-base & 200K & 32K \\
Legal-de-R-base & 200K & 32K \\
Legal-el-R-base & 200K & 32K \\
Legal-hu-R-base & 200K & 32K \\
Legal-ga-R-base & 200K & 32K \\
Legal-it-R-base & 200K & 32K \\
Legal-lv-R-base & 200K & 32K \\
Legal-lt-R-base & 200K & 32K \\
Legal-mt-R-base & 200K & 32K \\
Legal-pl-R-base & 200K & 32K \\
Legal-pt-R-base & 200K & 32K \\
Legal-ro-R-base & 200K & 32K \\
Legal-sk-R-base & 200K & 32K \\
Legal-sl-R-base & 200K & 32K \\
Legal-es-R-base & 200K & 32K \\
Legal-sv-R-base & 200K & 32K \\
Legal-XLM-R-base & 1M & 128K \\
Legal-XLM-R-large & 500K & 128K \\
Legal-XLM-LF-base & 50K & 128K \\
\bottomrule
\end{tabular}
}
\caption{Model Details}
\label{tab:model_details}
\end{table}

\onecolumn

\section{Hyperparameter Details}
\label{sec:hyperparameter_details}

\begin{table*}[!ht]
\fontsize{7pt}{7pt}\selectfont
\centering
\resizebox{\textwidth}{!}{
\begin{tabular}{lccccrrrrrrrr}
\toprule
       \bf{source} & \bf{Dataset} & \bf{Task} & \bf{Task type} &  \bf{Hierarchical} & \bf{Seeds} &  \bf{lower case} &  \bf{Batch size} & \bf{Metric for best model} & \bf{Evaluation strategy} & \bf{Epochs} &  \bf{Early stopping patience} &  \bf{Learning rate} \\
\midrule
\cite{niklaus_lextreme_2023} &          GLN &       GLN &            NER &              False &      1,2,3 &             True &               64 &            evaluation loss &                    epoch &          50 &                             5 &             1e-5 \\
\cite{niklaus_lextreme_2023} &          LNR &       LNR &            NER &              False &      1,2,3 &             True &               64 &            evaluation loss &                    epoch &          50 &                             5 &             1e-5 \\
\cite{niklaus_lextreme_2023} &          LNB &       LNB &            NER &              False &      1,2,3 &             True &               64 &            evaluation loss &                    epoch &          50 &                             5 &             1e-5 \\
\cite{niklaus_lextreme_2023} &          MAP &     MAP-F &            NER &              False &      1,2,3 &             True &               64 &            evaluation loss &                    epoch &          50 &                             5 &             1e-5 \\
\cite{niklaus_lextreme_2023} &          MAP &     MAP-C &            NER &              False &      1,2,3 &             True &               64 &            evaluation loss &                    epoch &          50 &                             5 &             1e-5 \\
\cite{niklaus_lextreme_2023} &          BCD &     BCD-J &           SLTC &               True &      1,2,3 &             True &               64 &            evaluation loss &                    epoch &          50 &                             5 &             1e-5 \\
\cite{niklaus_lextreme_2023} &          BCD &     BCD-U &           SLTC &               True &      1,2,3 &             True &               64 &            evaluation loss &                    epoch &          50 &                             5 &             1e-5 \\
\cite{niklaus_lextreme_2023} &          GAM &       GAM &           SLTC &              False &      1,2,3 &             True &               64 &            evaluation loss &                    epoch &          50 &                             5 &             1e-5 \\
\cite{niklaus_lextreme_2023} &          GLC &     GLC-C &           SLTC &               True &      1,2,3 &             True &               64 &            evaluation loss &                    epoch &          50 &                             5 &             1e-5 \\
\cite{niklaus_lextreme_2023} &          GLC &     GLC-S &           SLTC &               True &      1,2,3 &             True &               64 &            evaluation loss &                    epoch &          50 &                             5 &             1e-5 \\
\cite{niklaus_lextreme_2023} &          GLC &     GLC-V &           SLTC &               True &      1,2,3 &             True &               64 &            evaluation loss &                    epoch &          50 &                             5 &             1e-5 \\
\cite{niklaus_lextreme_2023} &          SJP &       SJP &           SLTC &               True &      1,2,3 &             True &               64 &            evaluation loss &                    epoch &          50 &                             5 &             1e-5 \\
\cite{niklaus_lextreme_2023} &          OTS &    OTS-UL &           SLTC &              False &      1,2,3 &             True &               64 &            evaluation loss &                    epoch &          50 &                             5 &             1e-5 \\
\cite{niklaus_lextreme_2023} &          OTS &    OTS-CT &           MLTC &              False &      1,2,3 &             True &               64 &            evaluation loss &                    epoch &          50 &                             5 &             1e-5 \\
\cite{niklaus_lextreme_2023} &          C19 &       C19 &           MLTC &              False &      1,2,3 &             True &               64 &            evaluation loss &                    epoch &          50 &                             5 &             1e-5 \\
\cite{niklaus_lextreme_2023} &          MEU &     MEU-1 &           MLTC &               True &      1,2,3 &             True &               64 &            evaluation loss &                          &             &                             5 &             1e-5 \\
\cite{niklaus_lextreme_2023} &          MEU &     MEU-2 &           MLTC &               True &      1,2,3 &             True &               64 &            evaluation loss &                          &             &                             5 &             1e-5 \\
\cite{niklaus_lextreme_2023} &          MEU &     MEU-3 &           MLTC &               True &      1,2,3 &             True &               64 &            evaluation loss &                          &             &                             5 &             1e-5 \\
\cite{chalkidis_lexglue_2022} &          ECtHR &      ECtHR-A &           MLTC &               True &  1,2,3,4,5 &             True &                8 &                   micro-f1 &                    epoch &          20 &                             3 &             3e-5 \\
\cite{chalkidis_lexglue_2022} &          ECtHR &      ECtHR-B &           MLTC &               True &  1,2,3,4,5 &             True &                8 &                   micro-f1 &                    epoch &          20 &                             3 &             3e-5 \\
\cite{chalkidis_lexglue_2022} &          EUR-LEX &       EUR-LEX &           MLTC &              False &  1,2,3,4,5 &             True &                8 &                   micro-f1 &                    epoch &          20 &                             3 &             3e-5 \\
\cite{chalkidis_lexglue_2022} &          SCOTUS &       SCOTUS &           SLTC &               True &  1,2,3,4,5 &             True &                8 &                   micro-f1 &                    epoch &          20 &                             3 &             3e-5 \\
\cite{chalkidis_lexglue_2022} &          LEDGAR &       LEDGAR &           SLTC &              False &  1,2,3,4,5 &             True &                8 &                   micro-f1 &                    epoch &          20 &                             3 &             3e-5 \\
\cite{chalkidis_lexglue_2022} &          UnfairToS &       UnfairToS &           MLTC &              False &  1,2,3,4,5 &             True &                8 &                   micro-f1 &                    epoch &          20 &                             3 &             3e-5 \\
\cite{chalkidis_lexglue_2022} &          CaseHOLD &       CaseHOLD &           MCQA &              False &  1,2,3,4,5 &             True &                8 &                   micro-f1 &                    epoch &          20 &                             3 &             3e-5 \\
\bottomrule
\end{tabular}

}
\caption{Hyperparameters for each dataset and task. However, there were a few exceptions. For the multilingual MEU tasks, given the dataset's size, we trained them for only 1 epoch with 1000 steps as the evaluation strategy when using multilingual models. When using monolingual models, we trained for 50 epochs with epoch-based evaluation strategy, as we utilized only the language-specific subset of the dataset. Regarding LexGlue, we followed the guidelines of \citet{chalkidis_lexglue_2022} for RoBERTa-based large language models, which required a maximum learning rate of 1e-5, a warm-up ratio of 0.1, and a weight decay rate of 0.06.}. 
\label{tab:hyperparameters}
\end{table*}

\section{Dataset Details}
\label{sec:dataset_details}

\begin{table*}[!ht]
\centering
\resizebox{\textwidth}{!}{
\begin{tabular}{ccccccccc}
\toprule
\bf{Language} & \bf{Text Type} & \bf{Words} & \bf{Documents} & \bf{Words per Document} & \bf{Jurisdiction} &                                                                                                                                                          \bf{Source} &                                  \bf{License/Copyright} \\
\midrule
\rowcolor{gray!25} \multicolumn{8}{c}{\fontsize{16}{18}\selectfont{\bf{Native Multi Legal Pile}}} \\

\midrule
           bg &    legislation  &       309M &           262k &                  1178 &          Bulgaria & \href{https://elrc-share.eu/repository/browse/marcell-bulgarian-legislative-subcorpus-v2/946267fe8d8711eb9c1a00155d026706d2c9267e5cdf4d75b5f02168f01906c6/                                                                                                                                }{MARCELL} &                                       CC0-1.0 \\
           \midrule
            &         &        &            &                   &           Czechia &                                                                 \href{https://lindat.mff.cuni.cz/repository/xmlui/handle/11372/LRT-3052}{CzCDC Constitutional Court} &                                  CC BY-NC 4.0 \\
           cs &        caselaw &       571M &           342k &                  1667 &           Czechia &                                                         \href{https://lindat.mff.cuni.cz/repository/xmlui/handle/11372/LRT-3052}{CzCDC Supreme Administrative Court} &                                  CC BY-NC 4.0 \\
            &        &        &            &                   &           Czechia &                                                                        \href{https://lindat.mff.cuni.cz/repository/xmlui/handle/11372/LRT-3052}{CzCDC Supreme Court} &                                  CC BY-NC 4.0 \\
           \midrule
           da &        caselaw &       211M &            92k &                  2275 &           Denmark &                                                                                                                             \href{https://huggingface.co/DDSC}{DDSC} & CC BY 4.0 and other, depending on the dataset \\
           \midrule
           da &    legislation &       653M &           296k &                  2201 &           Denmark &                                                                                                                             \href{https://huggingface.co/DDSC}{DDSC} & CC BY 4.0 and other, depending on the dataset \\
           \midrule
           de &        caselaw &      1786M &           614k &                  2905 &           Germany &                                                                                                                   \href{https://de.openlegaldata.io}{openlegaldata} &                                      ODbL-1.0 \\
            &         &       &            &                   &       Switzerland &                                                                                                                    \href{https://entscheidsuche.ch}{entscheidsuche} &                             similar to CC BY \\
           \midrule
           de &    legislation &       513M &           302k &                  1698 &           Germany &                                                                                                                   \href{https://de.openlegaldata.io}{openlegaldata} &                                      ODbL-1.0 \\
            &     &        &            &                   &       Switzerland &                                                                                                                  \href{https://www.lexfind.ch/fe/de/search}{lexfind} &                       not protected by copyright law \\
           \midrule
           en &    legislation &      2539M &           713k &                  3557 &       Switzerland &                                                                                                                  \href{https://www.lexfind.ch/fe/de/search}{lexfind} &                       not protected by copyright law \\
            &     &       &            &                   &                UK &                                                                                                                     \href{https://zenodo.org/record/6355465}{uk-lex} &                                     CC BY 4.0 \\
           \midrule
           fr &        caselaw &      1172M &           495k &                  2363 &           Belgium &                                                                                                                    \href{https://juportal.be/home/welkom}{jurportal} &                                  not protected by copyright law \\
            &         &       &            &                   &            France &                                                                                                            \href{https://echanges.dila.gouv.fr/OPENDATA/CASS}{CASS} &                              Open Licence 2.0 \\
            &         &       &            &                   &        Luxembourg &                                                                                                                      \href{https://justice.public.lu/fr.html}{judoc} &                               not protected by copyright law \\
            &         &       &            &                   &       Switzerland &                                                                                                                    \href{https://entscheidsuche.ch}{entscheidsuche} &                             similar to CC BY \\
           \midrule
           fr &    legislation &       600M &           253k &                  2365 &       Switzerland &                                                                                                                  \href{https://www.lexfind.ch/fe/fr/search}{lexfind} &                       not protected by copyright law \\
            &     &        &            &                   &           Belgium &                                                                                                    \href{https://www.ejustice.just.fgov.be/cgi/welcome.pl}{ejustice} &                       not protected by copyright law \\
           \midrule
           hu &    legislation &       265M &           259k &                  1019 &           Hungary & \href{https://elrc-share.eu/repository/browse/marcell-hungarian-legislative-subcorpus-v2/a87295ec8d6511eb9c1a00155d0267065f7e56dc7db34ce5aaae0b48a329daaa}{MARCELL} &                                       CC0-1.0 \\
           \midrule
           it &        caselaw &       407M &           159k &                  2554 &       Switzerland &                                                                                                                    \href{https://entscheidsuche.ch/}{entscheidsuche} &                             similar to CC BY \\
           \midrule
           it &    legislation &       543M &           238k &                  2278 &       Switzerland &                                                                                                                  \href{https://www.lexfind.ch/fe/it/search}{lexfind} &                       not protected by copyright law \\
           \midrule
           nl &    legislation &       551M &           243k &                  2263 &           Belgium &                                                                                                    \href{https://www.ejustice.just.fgov.be/cgi/welcome.pl}{ejustice} &                       not protected by copyright law \\
           \midrule
           pl &    legislation &       299M &           260k &                  1148 &            Poland &    \href{https://elrc-share.eu/repository/browse/marcell-polish-legislative-subcorpus-v2/dd14fa1c8d6811eb9c1a00155d026706c4718ddc9c6e4a92a88923816ca8b219}{MARCELL} &                                       CC0-1.0 \\
           \midrule
           pt &        caselaw &     12613M &            17M &                 728 &            Brazil &                                                                                                             \href{https://github.com/diego-feijo/rulingbr}{RulingBR} &                       not protected by copyright law \\
            &         &      &             &                  &            Brazil &                                                                                                      \href{https://www.kaggle.com/datasets/eliasjacob/brcad5}{CRETA} &                               CC BY-NC-SA 4.0 \\
            &         &      &             &                  &            Brazil &                                                                                                   \href{https://esaj.tjsp.jus.br/cjsg/consultaCompleta.do?f=1}{CJPG} &                       not protected by copyright law \\
           \midrule
           ro &    legislation &       559M &           396k &                  1410 &           Romania &  \href{https://elrc-share.eu/repository/browse/marcell-romanian-legislative-subcorpus-v2/2da548428b9d11eb9c1a00155d026706ce94a6b59ffc4b0e9fb5cd9cebe6889e}{MARCELL} &                                       CC0-1.0 \\
           \midrule
           sk &    legislation &       280M &           246k &                  1137 &          Slovakia &    \href{https://elrc-share.eu/repository/browse/marcell-slovak-legislative-subcorpus-v2/6bdee1d68c8311eb9c1a00155d0267063398d3f1a3af40e1b728468dcbd6efdd}{MARCELL} &                                       CC0-1.0 \\
           \midrule
           sl &    legislation &       366M &           257k &                  1418 &          Slovenia & \href{https://elrc-share.eu/repository/browse/marcell-slovenian-legislative-subcorpus-v2/e2a779868d4611eb9c1a00155d026706983c845a30d741b78e051faf91828b0d}{MARCELL} &                                     CC-BY-4.0 \\
           \midrule
        \bf total &                &    \bf 24236M &           \bf 23M &             \bf     1065 &                  &   \bf Native Multi Legal Pile \\
        \midrule
        \rowcolor{gray!25} \multicolumn{8}{c}{\fontsize{16}{18}\selectfont{\bf{Overall statistics for the remaining subsets}}} \\
\midrule
\bf total &                &    \bf 12107M  &           \bf 8M &             \bf     1457 &  EU               &    \bf Eurlex Resources                                                                                                                                                          &                \bf CC BY 4.0              \\
\midrule
\bf total &                &    \bf 43376M  &           \bf 18M &             \bf     2454 & US (99\%), Canada, and EU                &    \bf Pile of Law                                                                                                                                                          &             CC BY-NC-SA 4.0;    \bf See \citet{henderson_pile_nodate} for details             \\
\midrule
\bf total &                &    \bf 28599M   &           \bf 10M &             \bf     2454 &                  &    \bf Legal mC4                                                                                                                                                          &                \bf ODC-BY              \\
\bottomrule
\end{tabular}
}
\caption{Information about size and number of words and documents for \textit{Native} Multi Legal Pile are provided according to language and text type. For the remaining subsets of Multi Legal Pile we provide general statistics.}
\label{tab:multi_legal_pile}
\end{table*}

\end{document}